\documentclass[runningheads]{llncs}

\usepackage{eccv}

\usepackage{eccvabbrv}
\usepackage{xcolor}

\usepackage{graphicx}
\usepackage{booktabs}

\usepackage[accsupp]{axessibility}  % Improves PDF readability for those with disabilities.

\usepackage[pagebackref,breaklinks,colorlinks,citecolor=eccvblue]{hyperref}
\usepackage{orcidlink}

\begin{document}

%\title{Low Vision LLaVA: leveraging vision foundation model for visually impaired assistance} 
\title{A Light and Smart Wearable Platform with Multimodal Foundation Model for Enhanced Spatial Reasoning in People with Blindness and Low Vision}

\titlerunning{Smart Wearable for Enhanced Spatial Reasoning for Low Vision}

\author{Alexey Magay \and
Dhurba Tripathi \and
Yu Hao \and
Yi Fang}

\authorrunning{A.Magay et al.}

\institute{Embodied AI and Robotics (AIR) Lab\\
New York University Abu Dhabi, Abu Dhabi, UAE\\ 
\email{yfang@nyu.edu}\\
}

\maketitle

\begin{abstract}
   People with blindness and low vision (pBLV) face significant challenges, struggling to navigate environments and locate objects due to limited visual cues. Spatial reasoning is crucial for these individuals, as it enables them to understand and interpret the spatial relationships in their surroundings, enhancing their ability to navigate and interact more safely and independently. Current multi-modal large language (MLLM) models  for low vision people lack the spatial reasoning capabilities needed to effectively assist in these tasks. Moreover, there is a notable absence of lightweight, easy-to-use systems that allow pBLV to effectively perceive and interact with their surrounding environment. In this paper, we propose a novel spatial enhanced multi-modal large language model based approach for visually impaired individuals. By fine-tuning the MLLM to incorporate spatial reasoning capabilities, our method significantly improves the understanding of environmental context, which is critical for navigation and object recognition. The innovation extends to a hardware component, designed as an attachment for glasses, ensuring increased accessibility and ease of use. This integration leverages advanced VLMs to interpret visual data and provide real-time, spatially aware feedback to the user. Our approach aims to bridge the gap between advanced machine learning models and practical, user-friendly assistive devices, offering a robust solution for visually impaired users to navigate their surroundings more effectively and independently. The paper includes an in-depth evaluation using the VizWiz dataset, demonstrating substantial improvements in accuracy and user experience. Additionally, we design a comprehensive dataset to evaluate our method's effectiveness in real-world situations, demonstrating substantial improvements in accuracy and user experience.
\end{abstract}

\section{Introduction}
\label{sec:intro}

Visual impairment affects millions of people worldwide, significantly impacting their ability to perform everyday tasks independently. Global estimates highlight the urgent need for effective assistive technologies \cite{pascolini2012global}. Data from the World Health Organization underscores the increasing burden of visual impairment and blindness, emphasizing the necessity for innovative solutions \cite{world2014visual}. 

Multi-modal large language models \cite{2023gpt4, liu2024visual} hold significant potential for revolutionizing assistive technologies for the visually impaired. These models can process and interpret vast amounts of data to provide real-time, context-aware responses, making them particularly useful in question-answering systems for navigating and understanding complex environments \cite{yu2023reformulating}. Their ability to integrate and analyze diverse data types enhances their utility in creating systems that can interact naturally with users, offering them guidance and information tailored to their surroundings and immediate needs. An example of this is the VIAssist project, which adapts multi-modal large language models specifically for users with visual impairments, demonstrating significant improvements in accessibility and user interaction \cite{yang2024viassist}.

\begin{figure}[t]
    \centering
    \includegraphics[width=11.5cm]{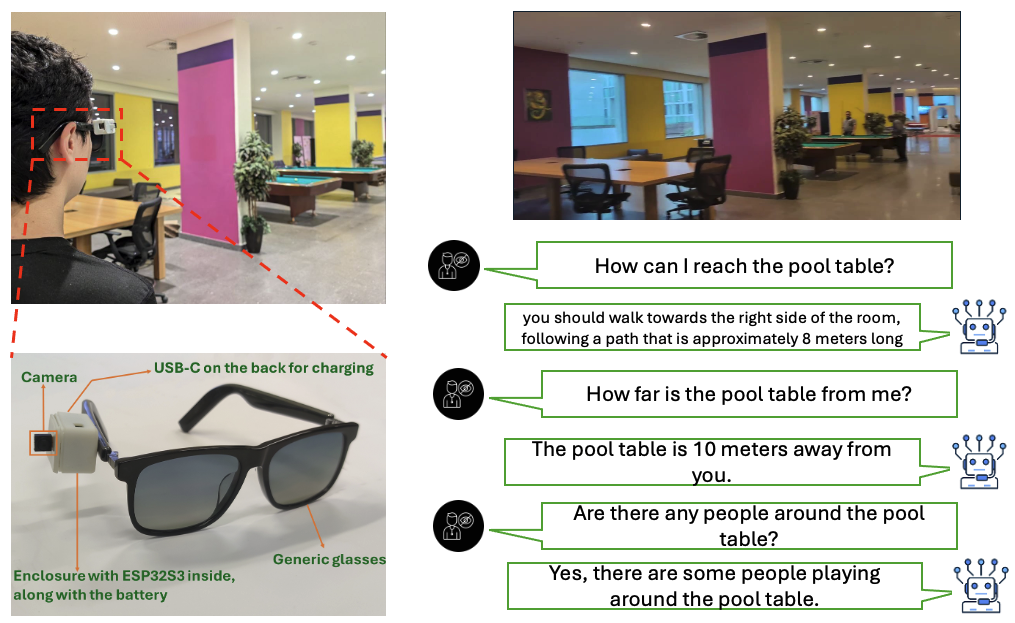}
    \caption{Overview of our proposed system: On the left, the lightweight and easy-to-use camera designed to be mounted on standard glasses. On the right, the fine-tuned multi-modal large language model (MLLM) enhanced with spatial reasoning capabilities for low vision assistance. }
    \label{fig6}
\end{figure}

However, while these technologies offer remarkable capabilities, they often lack effective spatial reasoning—a critical element for the visually impaired \cite{gui2019assistive}. Spatial reasoning enables users to comprehend and navigate their environment more effectively, allowing for enhanced independence. Existing technologies frequently fall short in delivering precise spatial information, which is essential for safe navigation and effective interaction with the environment. This limitation is particularly pronounced in systems that rely on sensory substitution, where the conversion of visual information into auditory or tactile cues may not convey accurate spatial relationships.

Recent advancements in mobile and wearable technologies have opened new avenues for assisting visually impaired individuals. Research on mobile assistive technologies demonstrates their potential to enhance the autonomy and mobility of visually impaired users by leveraging smartphones for real-time assistance. However, these technologies often face limitations in terms of battery life and integration with other devices \cite{hakobyan2013mobile}. Sensory substitution methods, which translate visual information into auditory or tactile cues, have shown promise in enhancing spatial navigation. One study explored stereosonic vision within a virtual reality paradigm, offering an immersive method for spatial navigation but faced challenges related to user adaptation and the complexity of virtual environments \cite{massiceti2018stereosonic}.
            
Wearable components can further enhance the effectiveness of visual-to-auditory sensory substitution systems. Devices such as smart glasses and haptic feedback systems provide continuous, hands-free assistance, improving user convenience. However, their adoption is often hindered by issues related to comfort, battery life, and the need for seamless integration with existing technologies \cite{fernandes2019review}.

%Understanding the principles of visual processing and attention is crucial for developing effective assistive technologies. The feature-integration theory of attention explains how visual features are processed and integrated into coherent perceptual experiences, informing the design of assistive systems that align with the natural cognitive processes of visually impaired users [6]. Studies on target search and identification performance in low vision patients reveal critical factors that must be addressed to enhance usability and efficiency in assistive technology design [7].

In this paper, as shown in Figure~\ref{fig6}, we propose a light and smart wearable platform integrated with a multimodal foundation model to enhance spatial reasoning for individuals with blindness and low vision. Specifically, we fine-tune the multimodal large language model to equip them with the ability to navigate and interact with environments, a task often compromised by the limited visual perception of depth, distance, and spatial relationships inherent in those with visual impairments. Furthermore, we introduce a lightweight, easy-to-use hardware component that can be attached to conventional glasses, enhancing the usability and accessibility of the technology. To demonstrate the effectiveness of our model's spatial reasoning capabilities, we have developed a specialized Low Vision Spatial Question Answering (LVSQA) dataset. Our real-world tests confirm the potential of our MLLM-based assistive technologies to provide precise navigation instructions and improved obstacle avoidance, thereby significantly enhancing user independence and experience.

Our contributions are summarized as follows:
 
\begin{enumerate}
\item We enhances multi-modal large language model capabilities by incorporating advanced spatial reasoning into assistive technologies. This addresses the limitation of existing solutions that often struggle with providing context-aware, precise navigation instructions for visually impaired users.
         
\item We introduce a novel wearable device that offers continuous, hands-free assistance. This integration overcomes the challenges of bulkiness, discomfort, and limited usability commonly associated with current wearable assistive technologies.

\item We propose the Low Vision Spatial Question Answering (LVSQA) dataset dataset for robust evaluation of the model's performance on visual question answering tasks that involve spatial reasoning. This dataset provides a comprehensive benchmark for assessing how well assistive technologies can handle spatial queries, thereby enhancing the development and fine-tuning of models to better serve visually impaired users.

\item We develop an integrated, lightweight, and smart wearable platform that combines enhanced multi-modal foundation models with advanced spatial reasoning capabilities, specifically designed for individuals with blindness and low vision.

\end{enumerate}

\section{Related Work}
\subsection{Foundation Models}

Foundation models, encompassing large-scale pretrained models such as Large Language Models (LLMs) and Multi-modal Large Language Models (MLLMs), have revolutionized the field of artificial intelligence by demonstrating exceptional capabilities across a wide range of tasks. These models, trained on vast amounts of data, are capable of understanding and generating human-like text, as well as interpreting visual inputs in a meaningful way. The GPT-3 model, for instance, has shown remarkable proficiency in natural language understanding and generation, enabling applications ranging from automated customer service to complex problem-solving \cite{brown2020language}. Similarly, MLLMs like CLIP \cite{radford2021learning} and VL-BERT \cite{su2019vl} have illustrated the potential of integrating visual and linguistic information, allowing for tasks such as image captioning and text-based image generation.

Recent advancements have introduced multimodal models that integrate capabilities across multiple modalities, further enhancing their utility. GPT-4 \cite{2023gpt4}, for instance, extends the capabilities of its predecessors by incorporating more sophisticated multimodal inputs, allowing it to process and generate text, images, and other forms of data simultaneously. This capability significantly enhances its application in fields requiring complex reasoning and contextual understanding across different types of information. LLaVA (Large Language and Vision Assistant) \cite{liu2024visual} is another recent model that exemplifies the integration of visual and linguistic processing. LLaVA is designed to handle tasks that require understanding and generating coherent responses based on both visual and textual inputs, such as detailed scene descriptions and interactive question-answering involving visual context.

Despite their impressive performance, foundation models face limitations, particularly in specialized domains requiring fine-grained understanding and reasoning. The black-box nature of these models often poses challenges in interpretability and reliability, especially in critical applications like assistive technology for visually impaired individuals. Furthermore, their generalist design may not adequately address specific needs such as spatial reasoning and real-time responsiveness, which are crucial for effective assistive technologies.

\subsection{Assistive Technology}
In recent years, numerous assistive technologies and applications have been developed to aid individuals with visual disabilities in comprehending their environment and improving their scene understanding \cite{boldini2020piezoelectric, giudice2008blind, massiceti2018stereosonic}. Traditional aids like white canes \cite{mcdaniel2008using} and guide dogs \cite{whitmarsh2005benefits} have been long-standing tools for enhancing mobility and spatial awareness. Technological advancements have further led to the creation of various assistive devices, including wearable cameras \cite{gupta2017indoor, rizzo2021covid, hao2022detect, hao2024multi}, GPS navigation systems, and object recognition technologies \cite{boldini2021inconspicuous}.

Wearable camera systems, such as OrCam MyEye and Seeing AI \cite{granquist2021evaluation}, provide real-time text reading and text-to-speech capabilities, delivering auditory feedback to individuals with visual impairments. These systems help with object identification, text reading, and facial recognition, thereby improving their interaction with their surroundings. GPS navigation systems like BlindSquare \cite{kumar2022study} and Lazarillo \cite{cardoso2019accessibility} use location-based services to offer audio instructions and navigation guidance for both indoor and outdoor environments.

Advancements in computer vision technologies have markedly enhanced scene comprehension capabilities for individuals with visual impairments. State-of-the-art object detection systems, leveraging deep learning architectures such as YOLO \cite{redmon2016you} and Faster R-CNN \cite{ren2016faster}, facilitate the real-time identification of objects within various environments. For instance, the Detect and Approach system employs YOLO to deliver a monocular-based navigation solution tailored for individuals with partial blindness and low vision, ensuring efficient and accurate object detection \cite{hao2022detect}. These integrations significantly improve accessibility and user interaction, thereby contributing to the field of assistive technology for the visually impaired.

Recent advancements in assistive technology for the visually impaired have also led to the development of various solutions addressing specific challenges. One approach integrates voice-based guidance with machine learning (ML) and deep learning (DL) algorithms into a wearable device, which includes navigation, face recognition, object detection, text-to-speech conversion, and currency recognition, all controlled via voice commands \cite{peraka2023novel, ahmed2022assistive}. Another development is an intelligent head-mounted obstacle avoidance device that focuses on real-time obstacle detection and warning, emphasizing computational efficiency and accommodating natural head movements \cite{xu2023intelligent}. Additionally, a wearable system using object detection, distance measurement, and tactile feedback supports navigation by providing real-time obstacle recognition and tactile cues through a glove with vibration patterns \cite{chen2023wearable}.

These studies demonstrate the application of ML and DL techniques in wearable devices to improve mobility, safety, and independence for visually impaired individuals. However, they also highlight the need for more integrated and context-aware solutions. Furthermore, the device designs presented in these works are often not user-friendly and may be challenging to integrate into real-world use due to their bulkiness and complexity. In contrast, our research aims to address these limitations by enhancing MLLMs with spatial reasoning capabilities and integrating them into a compact and powerful wearable device, making it more practical and accessible for everyday use.

\begin{figure}[t]
    \centering
    \includegraphics[width=12.5cm]{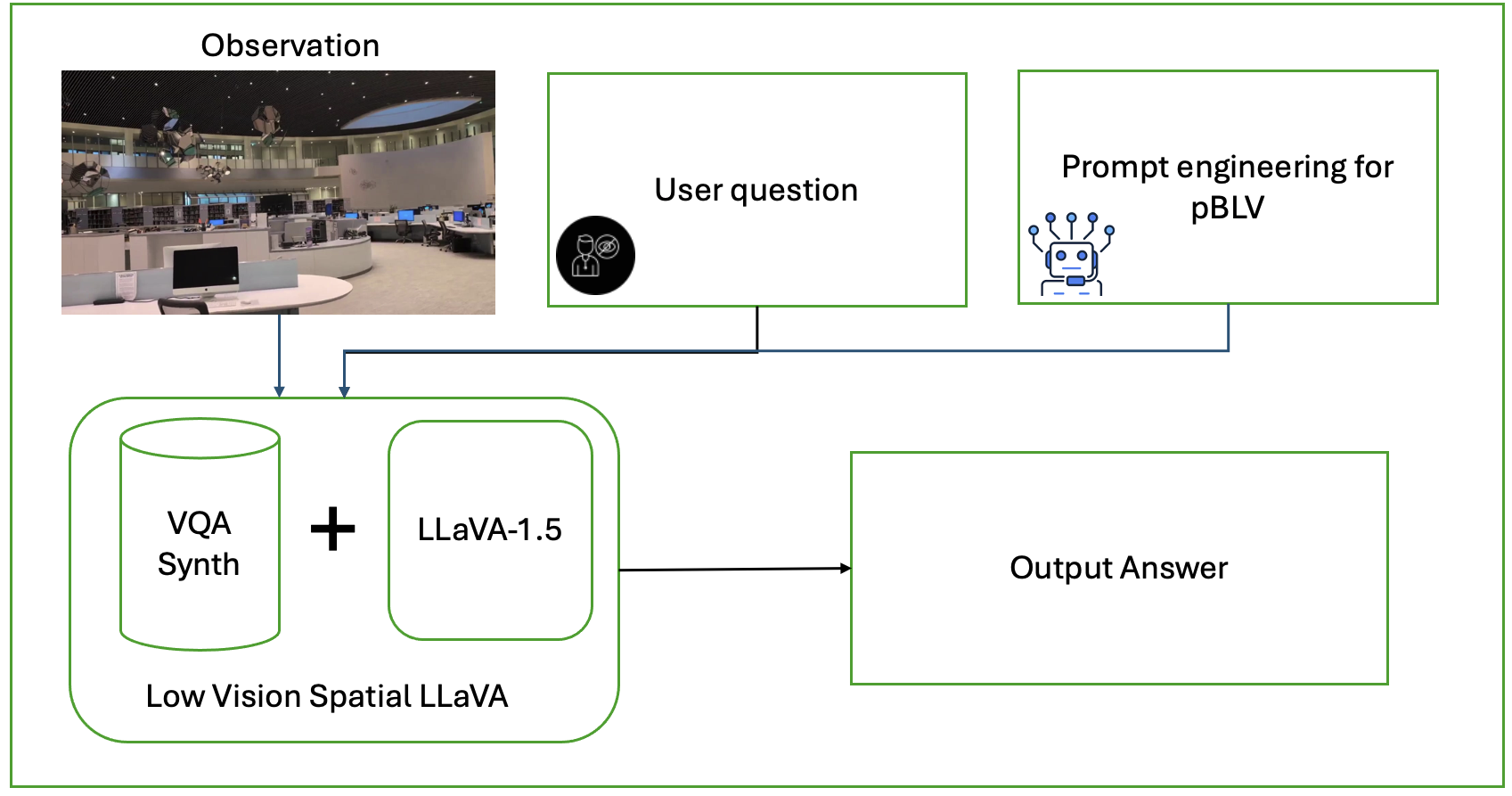}
    \caption{Flow of our proposed system. Given the observation captured by the camera and a user question, our proposed system use a fine-tuned Low Vision Spatial LLaVA model, which incorporates enhanced spatial reasoning capabilities. Together with specialized prompt engineering tailored for pBLV, the system generates comprehensive answers, effectively addressing the user's query based on surronding environment.}
    \label{fig0}
\end{figure}

\section{Method}
\subsection{Spatial Reasoning}
Spatial reasoning is crucial for individuals with low vision as it greatly enhances their ability to safely and effectively navigate and interact with their environment, which is often compromised by their limited ability to perceive visual cues like depth, distance, and spatial relationships \cite{green2023spatial}. This skill supports essential aspects of daily life, from navigating and moving independently through various settings, to accurately identifying the location and size of objects which aids in tasks such as identifying doorways and avoiding hazards. Moreover, enhanced spatial reasoning skills allow for more confident environmental interaction, facilitating complex spatial tasks like crossing streets and using public transportation, ultimately promoting greater independence and self-sufficiency for those with low vision.

In our method, we enhance the spatial reasoning capabilities of the Large Language and Vision Assistant (LLaVA) model \cite{liu2024visual} to improve assistive technology for visually impaired (VI) individuals. The method follows the framework proposed by \cite{chen2024spatialvlm} to fine-tune our Low Vision Spatial-LLaVA (LVS-LLaVA).

We start by filtering internet-scale images using a CLIP-based model to retain those that are suitable for spatial reasoning tasks. Pre-trained expert models extract object-centric contexts from these images, which are then converted into 3D point clouds using depth estimation techniques. This conversion allows us to capture accurate size and distance relationships, providing a comprehensive spatial context. To ensure precise object references, we use a user-configurable captioning approach that generates detailed and unambiguous descriptions.

The resulting Low Vision Spatial Question Answering (LVSQA) dataset includes a variety of qualitative and quantitative spatial reasoning questions. The LV-LLaVA model is trained on this dataset, integrating the spatial data with the original LLaVA training set. The training process is adjusted to emphasize spatial reasoning tasks, while still maintaining general VQA capabilities.

By training on this extensive LVSQA dataset, the low vision spatial LLaVA model achieves significant improvements in spatial reasoning, which is critical for assistive technology. It can accurately judge spatial relationships, provide precise metric estimations, and perform multi-step reasoning tasks. This enhanced spatial reasoning allows the technology to offer more accurate navigation instructions and better obstacle avoidance, improving the independence and safety of VI individuals. The integration of these capabilities into wearable devices ensures that users receive continuous, hands-free assistance in real-world environments, making the technology both practical and effective.

\begin{figure}[t]
    \centering
    \includegraphics[width=12.5cm]{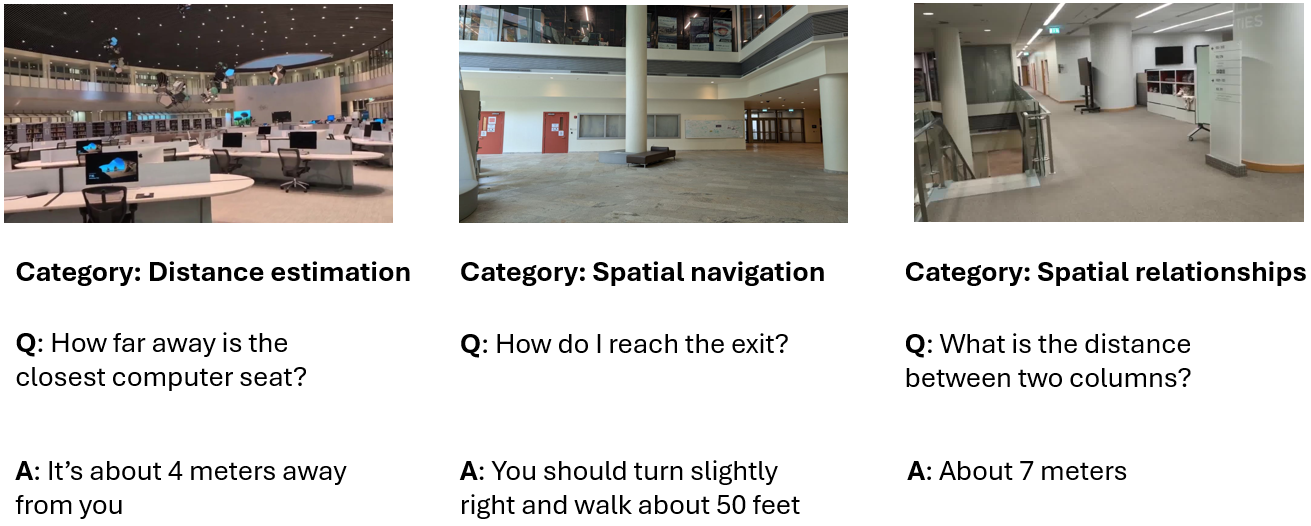}
    \caption{Examples from the proposed LVSQA dataset, featuring three categories: distance estimation, spatial navigation, and spatial relationships.}
    \label{fig1}
\end{figure}

\subsection{Low Vision Spatial Question Answering (LVSQA) Dataset Design}
The design of the Low Vision Spatial Question Answering dataset for this research is crafted to ensure a comprehensive and robust evaluation of spatial reasoning capabilities \cite{chen2024spatialvlm}, with a particular focus on aiding visually impaired and blind individuals. The dataset focuses on key objects of interest including exits/entrances, steps, elevators, hazards (pillars, trip hazards), seats, desks, and people. The process involves several critical steps.

The dataset begins with the selection of diverse images that are rich in spatial content and include the key objects of interest. These images are sourced from various environments to ensure variability and comprehensiveness, covering different indoor settings. Each image is manually annotated to identify the objects of interest, providing a solid foundation for generating relevant questions. The selection and annotation process is particularly focused on scenarios and objects that are crucial for navigation and safety for visually impaired and blind people.

For each annotated image, a set of questions is generated across three spatial reasoning categories. These categories include Navigational Guidance, Distance and Proximity, and Spatial Relationships. The questions are formulated using predefined templates tailored to elicit detailed spatial information, with a special emphasis on addressing the needs of visually impaired and blind individuals. Navigational Guidance questions aim at guiding navigation towards or around objects (e.g., ``\textit{How can I reach the [object]?}''). Distance and Proximity questions focus on the distance and reachability of objects (e.g., ``\textit{How far is the [object] from me?}''). Spatial Relationships questions concern the spatial relationships between multiple objects (e.g., ``\textit{Is the [object] above or below the [second object]?}''). The questions for each image are generated using the GPT-4 Vision Preview model, which analyzes the annotated objects in the image and applies the question templates to generate contextually relevant inquiries. This process ensures that the questions are directly applicable to real-world scenarios faced by visually impaired and blind individuals, enhancing their navigational assistance.

For each question generated, ground truth answers are manually created. These answers are based on detailed analysis and annotation of the images, ensuring high accuracy and reliability. The ground truths serve as a benchmark for evaluating the performance of VQA models on the dataset. The manual creation of ground truths is critical, as it ensures that the answers are tailored to the specific needs and safety considerations of visually impaired and blind users.

The final dataset is structured in a JSON file, where each entry includes the image file name as the key and an object containing question-answer pairs (one for each question category) generated for that image as the value.

\begin{figure}[t]
    \centering
    \includegraphics[width=11.5cm]{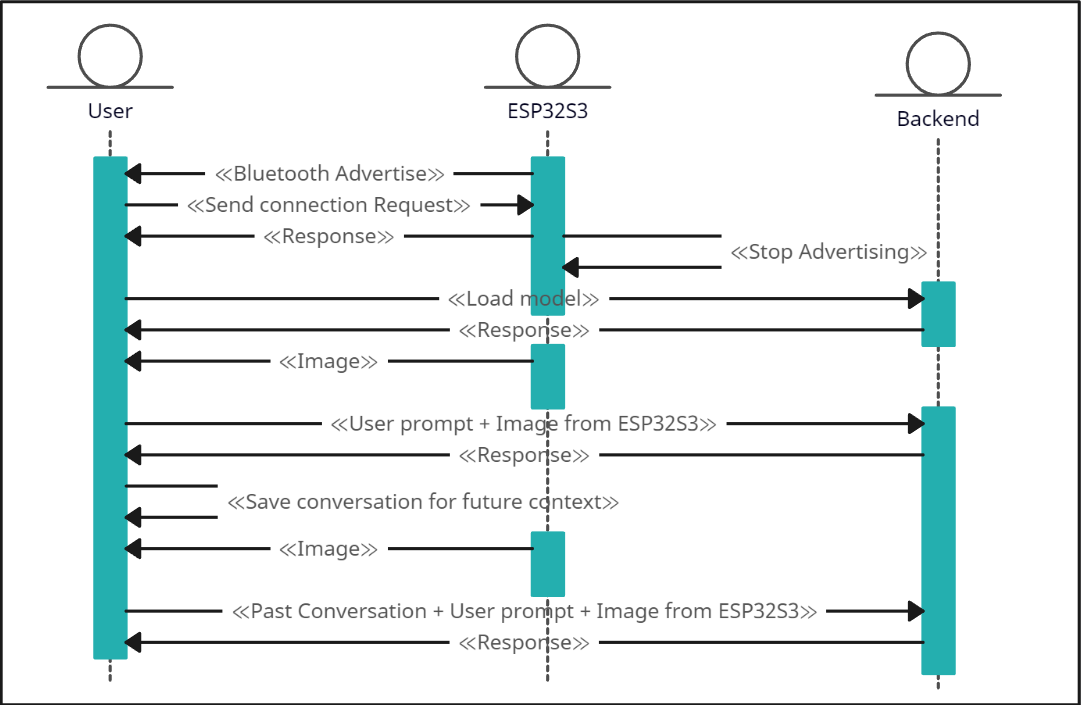}
    \caption{The overview of the system workflow. User Interaction diagram depicts how the user interacts with our system's backend through ESP32S3.}
    \label{fig3}
\end{figure}

\subsection{System Implementation Details}
In this section, we detail the implementation of our proposed system, which includes a wearable device designed to aid visually impaired users by providing real-time navigation and spatial awareness. The development of this device is driven by the goal of utilizing advancements in multi-modal large language models, integrated with wearable technology, to improve independence and enhance the quality of life for visually impaired individuals.

The hardware components of the system include a mobile phone tested on Android 15, the Seeed Studio XIAO ESP32 S3 Sense, and a server. The Seeed Studio XIAO ESP32 S3 Sense is a thumb-sized development board that integrates a camera sensor and Bluetooth for wireless communication, supports an SD card for image storage, and features the ESP32-S3R8 Xtensa LX7 dual-core, 32-bit processor operating up to 240 MHz, BLE Bluetooth 5.0 with Bluetooth mesh technology, and an OV2640 camera sensor for 1600x1200 resolution images \cite{johndoe_awesome_project_2023}. The compact design, powerful chip, and Arduino support make it ideal for this project. The server runs the model locally.

The system architecture consists of three major components: the image capture device, the frontend, and the backend. The OV2640 camera sensor mounted on the XIAO ESP32 S3 Sense uses BLE Bluetooth 5.0 to send images to the frontend. The Arduino code initializes BLE and advertises three services: a custom photo transmission service including a characteristic for photo data, a device information service, and a battery level service. The battery service updates and notifies the battery level every minute. When a BLE client connects, the camera captures JPEG images at XGA resolution every two seconds. Image data is sent in chunks of up to 182 bytes via BLE notifications, with 180 bytes for image data and 2 bytes for the frame index. An end flag marks the end of the transmission. The BLE server handles connection status changes and restarts advertising upon disconnection.

The frontend is developed using Expo, a React Native open-source framework for creating universal native apps for Android, iOS, and the web. The app features a minimalist, user-friendly design. On startup, it includes a connect button to scan for Bluetooth peripherals. The app uses the react-native-ble-plx library for BLE communication. Upon connection, a new interface is displayed. The app handles errors, permission requests, and Bluetooth service availability. It listens to the characteristic for photo data and reconstructs the original image from received chunks. Users can send requests to the backend via voice or text, using react-native-voice for voice recognition and expo-speech for text-to-speech. The app stores communication history for context in subsequent interactions and notifies the user if the connection is lost.

The backend is developed using Flask, a lightweight web framework for Python. It provides an API for interacting with the machine learning model and has four main endpoints: ``\textit{/load\_model}'' initializes and loads the model into memory using LLaVA15 ChatHandler and the Llama model from ``\textit{llama\_cpp}''; ``\textit{/process\_image}'' accepts POST requests with a base64-encoded image and a prompt, processing the image with the loaded model; ``\textit{/chat\_completion}'' handles POST requests with chat history; and 
``\textit{/close\_model}'' unloads the model from memory, freeing resources. The backend is designed to be efficient and scalable, running on port 54345 and handling multiple requests concurrently. The interaction with the wearable device is depicted in \ref{fig3}.

\begin{figure}[t]
    \centering
    \includegraphics[width=7.5cm]{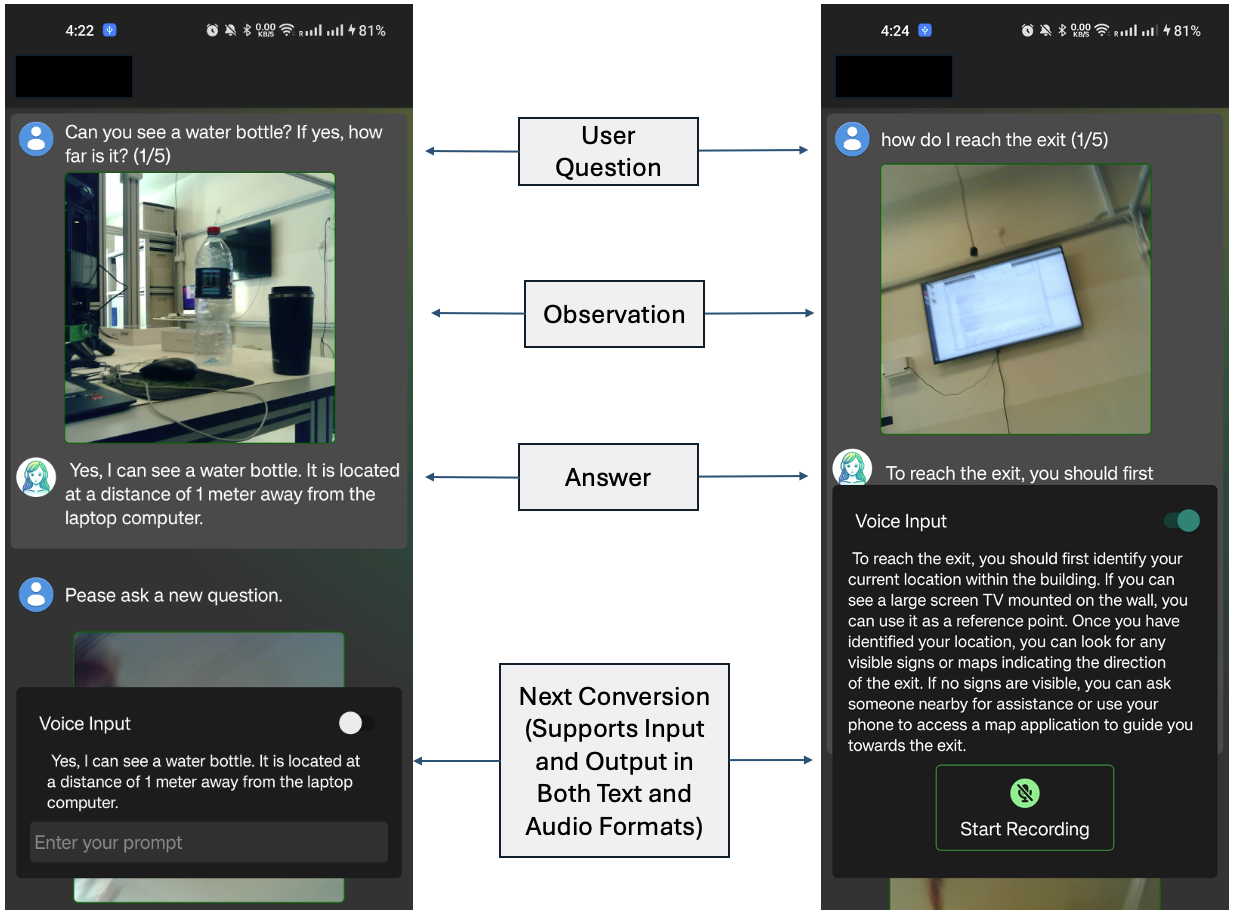}
    \caption{Examples of our system's smartphone application, demonstrating support for both text and audio formats in question and answer interactions.}
    \label{fig5}
\end{figure}

\section{Experiments}

\subsection{Experiments on LVSQA Dataset}

In this section, we conduct experiment on the proposed LVSQA dataset to evaluate the performance and usability of our assembled wearable device in providing navigation assistance and spatial awareness to visually impaired users. The method was tested with 100 observations (images) from various real-world indoor environments, including administrative buildings, public lounges, dining areas, and office spaces. A total of 300 user queries were made during these tests.

The performance of each model was evaluated using the following metrics:

\begin{itemize}
    \item \textbf{BLEU-1 and BLEU-2}: Measures the precision of n-grams (1-gram and 2-gram) in the generated responses compared to the reference answers \cite{papineni2002bleu}.
    \item \textbf{ROUGE}: Measures the recall of n-grams, providing insight into the coverage of the reference answers by the generated responses \cite{lin2004rouge}.
    \item \textbf{CIDEr}: Evaluates the consensus between the generated responses and the reference answers, focusing on the similarity of content \cite{vedantam2015cider}.
    \item \textbf{METEOR}: Considers both precision and recall, along with synonymy and stemming, to provide a more holistic evaluation of the generated responses \cite{banerjee2005meteor}.
\end{itemize}

The user queries were divided into three categories: distance estimation, object identification, and navigational questions. The results, summarized in Table \ref{tab2}, demonstrate the model's good spatial reasoning capabilities combined with inherently great question answering abilities. The model performed best at Distance Estimation, while the lowest performance was observed for Navigational questions. The primary reasons for this lower performance include the object mentioned in a question not being present in the image or being far away, and image blurriness.

\begin{table}[h]
    \centering
    \begin{tabular}{c|ccccc}
    \hline
         & BLEU-1 & BLEU-2 & ROUGE & CIDEr & METEOR\\
    \hline
        Navigation & 0.367 & 0.214 & 0.251 & 0.349 & 0.156\\
        Distance Estimation & 0.514 & 0.393 & 0.383 & 0.402 & 0.245\\
        Relationships & 0.418 & 0.302 & 0.320 & 0.415 & 0.224\\
    \hline
    \end{tabular}
    \caption{Performance of our model across three question categories in LVSQA dataset.}
    \label{tab2}
\end{table}

\begin{figure}
    \centering
    \includegraphics[width=11.5cm]{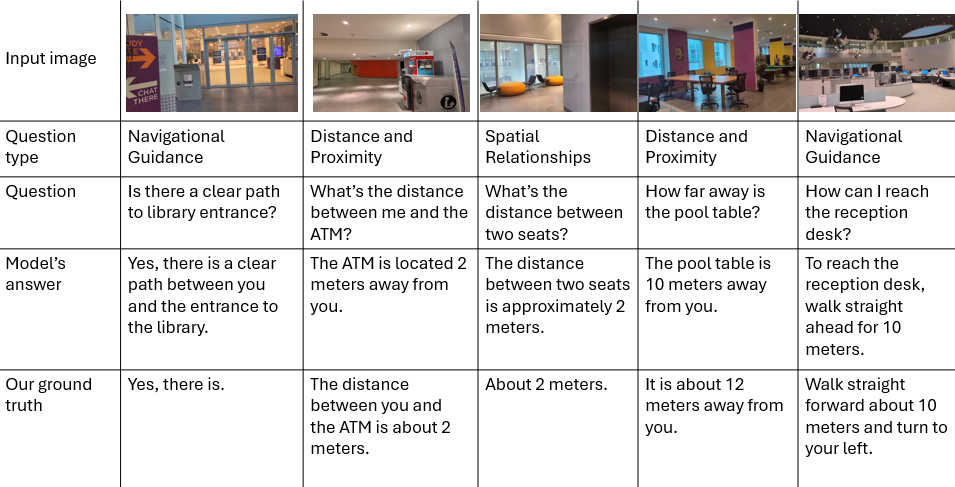}
    \caption{Qualitative results of the experiments for 5 random observations across all question categories in LVSQA dataset.}
    \label{fig4}
\end{figure}

In real-world scenarios, the majority of distance estimations made by the model in response to user queries fell within 5 meters. While these estimations were generally accurate, there is room for improvement to enhance precision. The table below provides the results of the experiments on our LVSQA dataset comprised of real-world indoor observations.

\subsection{Ablation Study}

To evaluate the effectiveness of our method in enhancing spatial reasoning tasks, we conducted an ablation study using a custom dataset. The metrics obtained from the experiments with Ours, GPT-4, and LLaVA are summarized in Table \ref{tab3}. 

\begin{itemize}
    \item \textbf{GPT-4}: A state-of-the-art language model known for its strong general reasoning abilities.
\item \textbf{LLaVA}: A Vision-Language model designed for visual question answering tasks.
\item \textbf{Ours}: Our fine-tuned Low Vision Spatial LLaVA model with enhanced spatial reasoning capabilities.
\end{itemize}

The results clearly demonstrate that our model outperforms both GPT-4 and LLaVA across all evaluated metrics, underscoring the improvements achieved through our approach.

\begin{table}[h]
    \centering
    \begin{tabular}{c|ccccc}
    \hline
         & BLEU-1 & BLEU-2 & ROUGE & CIDEr & METEOR\\
    \hline
        LLaVA & 0.387 & 0.298 & 0.343 & 0.391 & 0.193\\
        GPT-4 & 0.324 & 0.249 & 0.302 & 0.367 & 0.148\\
        Ours & 0.433 & 0.303 & 0.318 & 0.389 & 0.208\\
    \hline
    \end{tabular}
    \caption{Quantitative results of the ablation study on LVSQA dataset.}
    \label{tab3}
\end{table}

The significant increase in performance of the ours model can be attributed to two main factors. First, the model's enhanced spatial reasoning capabilities allow for better distance estimations, which are critical for accurately understanding and navigating environments. Second, the fine-tuning of the model specifically for spatial question answer tasks ensures that it is adept at interpreting and responding to complex queries about visual scenes. This combination of improved spatial reasoning and task-specific fine-tuning enables ours model to provide more precise and context-aware responses, leading to superior performance across all evaluated metrics.

These results from our ablation study demonstrate that the fine-tuning of the LLaVA model with spatial reasoning capabilities significantly enhances its performance in visual question answering tasks. The consistent improvement across all metrics highlights the effectiveness of our approach in providing more accurate and context-aware assistance, validating the potential of our model as a superior solution for spatial reasoning in assistive technologies.

\subsection{Experiments on VizWiz Dataset}

The VizWiz dataset \cite{bigham2010vizwiz}, known for its diverse and challenging visual question answering tasks, was used to evaluate the performance of the models. We conducted experiments with GPT-4, LLaVA, and our model. The results of our experiments are summarized in Table \ref{tab1}.

\begin{table}[h]
    \centering
    \begin{tabular}{c|ccccc}
    \hline
         & BLEU-1 & BLEU-2 & ROUGE & CIDEr & METEOR\\
    \hline
        GPT-4 & 0.480 & 0.293 & 0.326 & 0.412 & 0.169\\
        LLaVA & 0.650 & 0.396 & 0.401 & 0.424 & 0.205\\
        Ours & 0.618 & 0.415 & 0.425 & 0.407 & 0.210\\
    \hline
    \end{tabular}
    \caption{Quantitative results of experiments on VizWiz dataset.}
    \label{tab1}
\end{table}

The results indicate that our fine-tuned Low Vision Spatial LLaVA model performs comparably to GPT-4 and LLaVA across all evaluated metrics. Specifically, our model achieves a BLEU-1 score of 0.618, BLEU-2 score of 0.415, ROUGE score of 0.425, CIDEr score of 0.407, and METEOR score of 0.210. These results demonstrate that the addition of spatial reasoning through fine-tuning does not compromise the model's general reasoning ability.

The slight variations in scores across the models are within expected ranges, suggesting that our model maintains strong performance in general visual question answering tasks while offering enhanced spatial reasoning capabilities. This indicates that our approach successfully integrates spatial reasoning without detracting from the overall effectiveness of the model.

In conclusion, the experiments on the VizWiz dataset validate that the fine-tuning of LLaVA for spatial reasoning retains robust general reasoning abilities, confirming the efficacy of our enhancements in maintaining comprehensive model performance.

\section{Limitations and Future Work}

\subsection{Limitations}

In our experiments, a more qualitative approach was chosen to evaluate the performance of our model. This decision was based on the understanding that humans do not require the high level of precision necessary for robotic applications. The focus was on the model's ability to provide useful and context-aware navigation and spatial awareness instructions to visually impaired users. However, this qualitative approach has its limitations, particularly in tasks such as distance estimation where a more quantitative evaluation is essential. Future work should include developing and implementing a rigorous quantitative framework to assess the model's accuracy in distance estimation and other spatial reasoning tasks.

\subsection{Future Work}

Building on the current research, several avenues for future work can be pursued to enhance our model. Improving distance estimation accuracy is crucial, and this can be achieved by implementing quantitative metrics to evaluate and refine the model's performance in estimating distances and spatial relationships. This will involve creating a benchmark dataset specifically for distance estimation and conducting detailed analysis. Additionally, enhancing the user interface and experience is vital. Improving the Android application's user interface to provide a more intuitive and seamless experience involves incorporating user feedback to refine the auditory feedback system and ensuring that the device is comfortable for extended use.

\section{Conclusion}

In this paper, we presented the development and evaluation of our model, a multi-modal large language model enhanced with spatial reasoning capabilities, aimed at improving assistive technology for visually impaired individuals. Our approach involved fine-tuning the LLaVA model using a Low Vision Spatial Question Answer dataset and integrating it into a wearable device paired with an Android application.

Through real-world tests conducted in various indoor environments, we demonstrated that our model significantly enhances navigation and spatial awareness for visually impaired users. The model's ability to provide accurate and context-aware instructions was validated, showing that the addition of spatial reasoning capabilities does not compromise the general reasoning ability of the model. Our experiments on the VizWiz dataset further confirmed this, with our model performing comparably to state-of-the-art models such as GPT-4 and LLaVA across multiple evaluation metrics.

Despite the successes, several challenges were encountered during the research and prototype development, which will inform future improvements. Additionally, our qualitative evaluation approach highlighted the need for more quantitative assessment methods, particularly for tasks such as distance estimation.

Future work will focus on addressing the identified challenges, enhancing the accuracy of distance estimation, expanding real-world testing, improving the user interface, exploring multimodal integration, and ensuring the scalability and deployment of our model for broader use. By pursuing these avenues, we aim to provide a more comprehensive and reliable assistive solution that significantly improves the independence and quality of life for visually impaired individuals.

\clearpage  % TODO REVIEW/FINAL: This \clearpage needs to be removed from both review and camera-ready versions.

\bibliographystyle{splncs04}
\bibliography{main}
\end{document}